\documentclass[final,5p,times,twocolumn]{elsarticle}

\usepackage{amssymb}
\usepackage{amsmath}

\usepackage{booktabs}
\usepackage{subcaption}
\usepackage{xcolor}
\usepackage{makecell}

\newcommand{\matx}[1]{\mathbf{#1}}
\newcommand{\vect}[1]{\mathbf{#1}}
\newcommand{\norm}[1]{\Vert #1 \Vert}
\newcommand{\Norm}[1]{\left\Vert #1 \right\Vert}

\newcommand{\better}[1]{\textcolor{blue}{#1}}
\newcommand{\worse}[1]{\textcolor{red}{#1}}

\DeclareMathOperator{\argmax}{arg max}

\journal{Pattern Recognition Letters}

\begin{document}

\begin{frontmatter}



\title{An empirical evaluation of rewiring approaches in graph neural networks} 


\author{Alessio Micheli} 
\ead{alessio.micheli@unipi.it}

\author{Domenico Tortorella\corref{cor1}}
\ead{domenico.tortorella@unipi.it}
\cortext[cor1]{Corresponding author}

\affiliation{organization={Department of Computer Science, University of Pisa},
            addressline={largo Bruno Pontecorvo, 3}, 
            city={Pisa},
            postcode={56127}, 
            country={Italy}}

\begin{abstract}
Graph neural networks compute node representations by performing multiple message-passing steps that consist in local aggregations of node features.
Having deep models that can leverage longer-range interactions between nodes is hindered by the issues of over-smoothing and over-squashing.
In particular, the latter is attributed to the graph topology which guides the message-passing, causing a node representation to become insensitive to information contained at distant nodes.
Many graph rewiring methods have been proposed to remedy or mitigate this problem.
However, properly evaluating the benefits of these methods is made difficult by the coupling of over-squashing with other issues strictly related to model training, such as vanishing gradients.
Therefore, we propose an evaluation setting based on message-passing models that do not require training to compute node and graph representations.
We perform a systematic experimental comparison on real-world node and graph classification tasks, showing that rewiring the underlying graph rarely does confer a practical benefit for message-passing.
\end{abstract}


\end{frontmatter}



\section{Introduction}
\label{sec:intro}

Neural models for graphs \cite{Bacciu2020,Wu2021}, commonly also called \emph{graph neural networks} (GNNs), have been successfully applied in many real-world tasks, such as identifying categories of users in social networks or classifying molecules.
GNNs typically operate in the \emph{message-passing} paradigm, that is by exchanging information between nearby nodes according to the graph structure.
Messages are computed from the neighbor node features, then aggregated by a permutation-invariant function to provide node representations.
With multiple message-passing steps, GNNs are able to learn a hierarchy of representations that capture interactions between increasingly distant nodes.
This is accomplished either via multiple iterations of the same parameterized message-passing function \cite{Scarselli2009,Gallicchio2010}, or by a deep network of message-passing layers with different learnable parameters (proposed originally in \cite{Micheli2009} and subsequently in \cite{Duvenaud2015,Atwood2016,Kipf2017}).
The need for sufficiently deep graph networks arises for tasks that require the discovery of long-range dependencies between nodes, otherwise the model incurs in \emph{under-reaching} \cite{Alon2021}.
As deep learning on graphs progressed, several challenges preventing the computation of effective node representations have emerged.
Among those, \emph{over-squashing} is inherently connected to the inductive bias at the base of GNNs: the problem of encoding an exponentially growing receptive field \cite{Micheli2009} in a fixed-size node embedding dimension \cite{Alon2021}.
As simply increasing the width of node representations does not remove the underlying issues caused by the graph topology \cite{DiGiovanni2023}, this has motivated a growing number of methods that alter (i.e. \emph{rewire}) the original graph as a pre-processing to improve message-passing.
In this paper, we attempt to meet the need for an \emph{empirical approach} to assess the benefits of graph rewiring methods.
Indeed, altering the input data without taking into account the specific learning task can possibly lead to the loss of critical information.
Since the quality of node representations computed on rewired graphs is evaluated according to the accuracy in downstream learning tasks, the use of end-to-end trained models does not allow for decoupling the effects caused by graph topology on message-passing from the problems inherently connected to training in deep neural networks.
Indeed, while it has been proven that gradient vanishing prevails on over-squashing when the number of message-passing steps is much larger than the range of node interactions needed to solve the task, it is still unclear how the two issues interact with each other or what happens in intermediate regimes \cite{DiGiovanni2023}.
Furthermore, GNN models that completely or partially avoid learning representations via training have exhibited performances close to or above common end-to-end trained ones \cite{Gallicchio2020aaai,Gallicchio2020ijcnn,Tortorella2023neurocomp,Huang2023}, in particular when compared to previous results for rewiring methods applied to trained GNNs \cite{Tortorella2022log}.
Therefore, as opposed to previous literature, we propose to use message-passing models that compute node representations \emph{without training}, either by being parameter-free \cite{Wu2019} or by following the reservoir computing paradigm \cite{Gallicchio2010}, where parameters are just randomly initialized under certain constraints.
Crucially, the issues that graph rewiring methods aim to address are connected with the inductive bias of GNNs \cite{Battaglia2018}, that is to the message-passing \emph{per se}, whether is done in the forward or backward pass.
This will allow us to assess the actual benefits of graph rewiring on several node and graph classification tasks.

In the following Sec.~\ref{sec:rewiring}, we present a brief survey of the rewiring methods that will be evaluated in our experiments.
In Sec.~\ref{sec:gnns}, we introduce SGC and GESN, the two training-free message-passing models adopted in our experimental framework.
The datasets and results of our experiments will be discussed in Sec.~\ref{sec:exper}, focusing our analysis on the effects of rewiring both on classification accuracy as well as on altering the graph topology.
Final conclusions are drawn in Sec.~\ref{sec:concl}.

\section{Graph rewiring methods}
\label{sec:rewiring}

Let $\mathcal{G}(\mathcal{V}, \mathcal{E})$ be a graph with nodes $v \in \mathcal{V}$ and edges $(u, v) \in \mathcal{E}$, each node having associated input features $\vect{x}_v \in \mathbb{R}^X$.
We denote by $\mathcal{N}_v$ the set of neighbors of node $v$ with cardinality (i.e. degree) $d_v$, and respectively by $\matx{A}$, $\matx{D}$, $\matx{L}$ the graph adjacency, degree and Laplacian matrices.
We also define the symmetric normalized adjacency $\matx{A}_\text{sym} = \matx{D}^{-\frac{1}{2}} \matx{A} \matx{D}^{-\frac{1}{2}}$, the random-walk normalized adjacency $\matx{A}_\text{rw} = \matx{A} \matx{D}^{-1}$, and the mean-aggregation normalized adjacency $\matx{A}_\text{mean} = \matx{D}^{-1} \matx{A}$, along with the respective normalized Laplacians $\matx{L}_\text{sym}$, $\matx{L}_\text{rw}$, $\matx{L}_\text{mean}$, and the self-loop augmented $\matx{\hat{A}} = \matx{A} + \matx{I}$.
Finally, we denote by $\matx{A}^+$ the pseudo-inverse of matrix $\matx{A}$.
Throughout the paper, we assume the graphs to be undirected.

\begin{figure*}
	\centering
	\includegraphics[width=.9\textwidth]{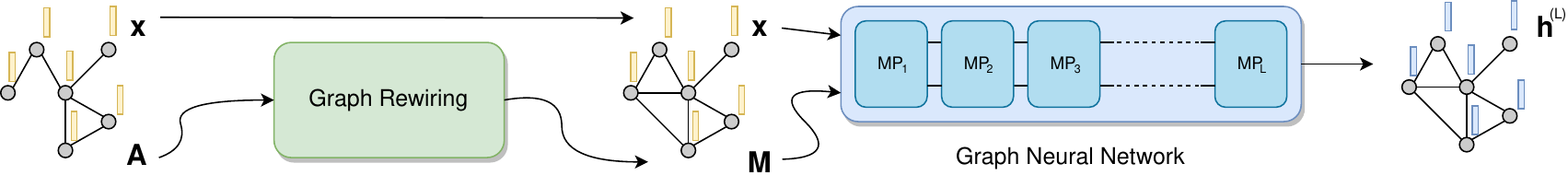}
	\caption{Graph rewiring as a pre-processing step: the rewiring algorithm changes the original graph connectivity $\matx{A}$ into a new connectivity matrix $\matx{M}$ that is then employed for message passing in a deep $L$-layer graph neural network to produce the node embeddings $\vect{h}^{(L)}$ for the downstream task.}
	\label{fig:rewiring-gnn}
\end{figure*}

A graph neural network (GNN) computes node representations $\vect{h}_v \in \mathbb{R}^H$ via a deep neural network of $L$ message-passing layers.
Each layer $k = 1, ..., L$ computes a new node representation $\vect{h}^{(k)}_v$ by performing a permutation-invariant aggregation of messages computed from the previous layer representations of neighbor nodes $\vect{h}^{(k-1)}_u$.
Without much loss of generality, we assume the message-passing layers to have the form
\begin{equation}\label{eq:mpnn}\textstyle
	\vect{h}^{(k)}_v = \phi_k \left( \sum_{u \in \mathcal{V}} M_{v u}\, \psi_k \left(\vect{h}^{(k-1)}_u\right) \right), \quad \vect{h}^{(0)}_v = \vect{x}_v,
\end{equation}
where local neighbors of $v$ are implicitly defined as nodes $u$ such that $M_{v u} \neq 0$.
By $\matx{M}$ we denote the message-passing matrix, which is a graph shift operator.
Such operator that can be e.g. either the adjacency $\matx{A}$, the Laplacian $\matx{L}$, or one of their normalizations.
In this case, the aggregations are performed on graph neighborhoods $\mathcal{N}_v$.
Message-passing layers can thus represent the relationships induced by graph connectivity in an efficient manner by leveraging the graph structure sparsity.
To capture long-range dependencies between nodes, GNNs must perform at least as many message-passing steps (i.e., have as many layers) as the distance between node pairs to avoid under-reaching \cite{Alon2021}.
However, building deep GNNs presents an inherent challenge.
As depth increases, the receptive field of nodes \cite{Micheli2009} grows exponentially, thus requiring more information to be encoded in the same fixed-size vectors.
This problem is called over-squashing \cite{Alon2021}.
Topping et al. \cite{Topping2022} have investigated this phenomenon via the analysis of node representations' sensitivity to input features.
Assuming there exist an $L$-path from node $u$ to node $v$, the sensitivity of $\vect{h}^{(L)}_v$ to input features $\vect{x}_u$ is upper bounded by
\begin{equation}\label{eq:sensitivity}
	\Norm{\frac{\partial \vect{h}^{(L)}_v}{\partial \vect{x}_u}} \;\leq\; \underbrace{\prod_{k=1}^{L} \norm{\phi_k} \norm{\psi_k}}_{\text{Lipschitz constants}}\; \underbrace{\left(\matx{M}^L\right)_{v u}}_{\text{graph topology}}.
\end{equation}
Over-squashing arises when the derivative in \eqref{eq:sensitivity} becomes too small, indicating that the representation of node $v$ is mostly insensitive to the information initially present at node $u$.
While increasing the layer Lipschitz constants or the dimension $H$ can mitigate the issue \cite{Tortorella2022log,DiGiovanni2023}, this may come at the expense of model generalization \cite{Maskey2022}.
Therefore, different methods have been proposed to alter the graph topology in a more favorable way to message-passing.
In this paper, we focus on graph rewiring methods that change the initial graph---or equivalently, the message-passing matrix $\matx{M}$---as a pre-processing step (Fig.~\ref{fig:rewiring-gnn}), as opposed to e.g. the implicit rewiring done by attention mechanisms \cite{Velickovic2018}.

\subsection{Diffusion processes}
\label{sec:diffusion}
Graph diffusion was originally proposed as a way of aggregating nodes beyond the immediate $1$-hop neighborhood \cite{Klicpera2019}, thus allowing a single message-passing layer to consider information from more distant nodes directly.
The generalized graph diffusion matrix is computed by the power series $\sum_{m = 0}^{\infty} \theta_m \matx{A}^m$, where the choice of coefficients $\theta_m$ defines the particular diffusion process and $\matx{A}$ can be replaced by any other transition matrix.
Two examples of graph diffusion are the heat kernel \cite{Kondor2002} with $\theta_m^\text{Heat} = e^{-t} \frac{1}{m!}, t > 0$, and personalized PageRank \cite{Page1999} with $\theta_m^\text{PageRank} = \alpha (1 - \alpha)^m, 0 < \alpha < 1$, which correspond respectively to the message-passing matrices
\begin{eqnarray}\label{eq:diffusion}
	\matx{M}_\text{Heat} = e^{-t \matx{A}} &\text{and}& \matx{M}_\text{PageRank} = \alpha (\matx{I} - (1 - \alpha) \matx{A})^+.
\end{eqnarray}
Diffusion-based rewiring was proposed exclusively for node-level tasks.

\subsection{Local bottlenecks}
\label{sec:sdrf}
In their analysis of over-squashing, Topping et al. \cite{Topping2022} have linked its causes to \emph{bottlenecks} in the graph topology that happen where the graph structure locally resembles a tree.
Intuitively, for a tree, the receptive field grows exponentially in the branching factor, while at the other opposite, a complete graph has a constant receptive field.
To provide a metric to quantify this local behavior, they have proposed the balanced Forman curvature, defined as
\begin{equation}\label{eq:ricci}\small
	\mathsf{Ric}_{u v} = \underbrace{\frac{2}{d_u} + \frac{2}{d_v}}_{\text{tree-likeness}} - 2 + \underbrace{2 \frac{\sharp^\triangle_{u v}}{\max\{d_u, d_v\}} + \frac{\sharp^\triangle_{u v}}{\min\{d_u, d_v\}}}_{\text{local similarity to a complete graph}} + \underbrace{\frac{\sharp^\square_{u v} + \sharp^\square_{v u}}{\gamma^\text{max}_{u v} \max\{d_u, d_v\}}}_{\text{grid-likeness}},
\end{equation}
where $\sharp^\triangle_{u v}$ is the number of triangles on the edge $(u, v)$, $\sharp^\square_{u v}$ is the number of neighbors of $u$ forming a $4$-cycle based on the edge $(u, v)$ without diagonals inside, and $\gamma^\text{max}_{u v}$ is a normalization factor.
For a graph having only positively curved edges (i.e. $\mathsf{Ric}_{u v} > 0$ for all $(u, v)$) it has been proved that the receptive field grows at most polynomially \cite{Topping2022}.
Therefore, rewiring algorithms that aim at increasing the graph curvature have been proposed.
\textsl{Stochastic Discrete Ricci Flow} (SDRF) \cite{Topping2022} iteratively samples an edge $(u, v)$ proportionally to how negatively curved it is, then adds the new edge $(u', v')$ able to provide the largest increase of $\mathsf{Ric}_{u v}$.
(The algorithm optionally removes the most positively curved edges to avoid growing the graph excessively.)

\subsection{Global bottlenecks}
\label{sec:grlef}
The edge curvature defined in equation \eqref{eq:ricci} is not the only way to measure the presence of bottlenecks in the graph topology.
A more \emph{global} metric is the Cheeger constant $\mathfrak{h}_\mathcal{G}$, which quantifies the minimum fraction of edges that need to be removed in order to make the graph disconnected.
A small $\mathfrak{h}_\mathcal{G}$ thus indicates that few edges act as a bridge between two otherwise disconnected communities.
However, computing the Cheeger constant is an NP-hard problem, so the lower bound given by the spectral gap $\lambda_1$ (i.e. the smallest positive Laplacian eigenvalue) is used as a proxy measure in practice: $\mathfrak{h}_\mathcal{G} \geq \frac{1}{2} \lambda_1$ \cite{Mohar1989}.
\textsl{Greedy Random Local Edge Flip} (GRLEF) \cite{Banerjee2022} proposes to improve a graph spectral gap by working exclusively \emph{locally} via the triangle counts $\sharp^\triangle_{u v}$, which are cheaper to compute as they require just neighborhood information.
The algorithm iteratively samples an edge $(u, v)$ proportionally to the inverse of triangle count, that is, from an area of the graph that is locally far away from being fully connected.
Then it chooses the pair of edges $(u, u'), (v, v')$ to flip into $(u, v'), (v, u')$ which provides the smallest net change in triangle count. 
This behavior can be interpreted as mitigating a very low local curvature (as suggested by the small term $\sharp^\triangle_{u v}$ in $\mathsf{Ric}_{u v}$) at the expense of a reduction in curvature of more positively curved neighboring edges.
\citet{Banerjee2022} supported the approach of their rewiring algorithm by empirically finding a correspondence between triangle count decrease and spectral gap increase.

\subsection{Expander propagation}
\label{sec:egp}
There is a class of graphs that avoid global bottlenecks by construction: expander graphs are simultaneously sparse and highly connected \cite{Hoory2006}.
Additionally, expander families of graphs are characterized by a uniform lower bound on the Cheeger constant \cite{Alon1986}, and for uniform maximal node degree, their diameter is also logarithmic in the number of nodes \cite{Mohar1991,Alon1985}.
\citet{Deac2022} have thus proposed to interleave the message propagation on the original graph with message-passing on an expander graph to provide for information propagation over bottlenecks.
The expander graphs adopted for \textsl{Expander Graph Propagation} (EGP) \cite{Deac2022} are derived from the Cayley graphs of finite groups $\mathrm{SL}(2, \mathbb{Z}_n)$, which are $4$-regular and thus guarantee sparsity.
Interestingly, these graphs have all negatively curved edges with $\mathsf{Ric}_{u v} = -\frac{1}{2}$.
In our experiments, we will thus use the message-passing matrix $\matx{M}_\text{EGP} = \matx{A}_\text{Cay}\, \matx{A}$, where $\matx{A}_\text{Cay}$ is the adjacency matrix of said Cayley graphs.

\subsection{Effective resistance}
\label{sec:diffwire}
Effective resistance \cite{Ellens2011} provides an additional way to measure bottlenecks in graph topology.
The resistance $\mathsf{Res}_{u v}$ between two nodes is proportional to the commute time $\mathsf{Com}_{u v}$, which is the number of expected steps for a random walk to go back and forth between nodes $u, v$.
A high resistance between two nodes is an indication of the difficulty for messages to pass from node $u$ to node $v$.
Black et al. \cite{Black2023} proved a sensitivity bound similar to \eqref{eq:sensitivity} relating high effective resistance $\mathsf{Res}_{v u}$ between pairs of nodes to a reduced sensitivity of the representations $\vect{h}_v^{(L)}$ to input features $\vect{x}_u$.
Furthermore, effective resistance is inversely related to the square of the Cheeger constant by the inequality $\max_{(u,v) \in \mathcal{E}} \mathsf{Res}_{u v} \leq \frac{1}{\mathfrak{h}_\mathcal{G}^2}$ \cite{ArnaizRodriguez2022}.
\citet{ArnaizRodriguez2022} have proposed a layer for learning effective resistance to re-weight the original graph adjacency (hence `\textsl{DiffWire}') in the perspective of sampling a spectrally similar but sparser graph which preserves the graph structural information \cite{Spielman2011}.
The additional intuitive effect is to enlarge the relative capacity of high-resistance edges, which correspond to bridges over more densely connected communities.
In our experiments, we implement the DiffWire approach by computing the effective resistance in exact form by $\mathsf{Res}_{u v} = (\vect{1}_u - \vect{1}_v)^\top \matx{L}^+ (\vect{1}_u - \vect{1}_v)$ with $\vect{1}_u$ the indicator vector of node $u$.
The resulting message-passing matrix therefore is $\matx{M}_\text{DiffWire} = \mathsf{Res} \odot \matx{A}$, where `$\odot$' denotes the elementwise product.

\section{Training-free graph neural networks}
\label{sec:gnns}

Since graph rewiring methods work as a pre-processing step on the input graph, the choice of the GNN model is crucial to assess their benefits in downstream task accuracy.
So far, only end-to-end trained models have been used, such as \textsl{Graph Convolution Network} (GCN) \cite{Kipf2017} in \cite{Topping2022}.
This approach does not allow for considering the effects of over-squashing independently from the other issues that can affect training in message-passing models, such as gradient vanishing.
By learning node and graph representations jointly with the task prediction readout, the experimental results become inextricably linked to how training is conducted.
Therefore, we propose to apply GNNs that compute node and graph representation \emph{without training} in our experimental setting for assessing the actual contributions of graph rewiring.
Indeed, rewiring methods aim to address issues connected with the model bias itself, that is local aggregation of messages computed from node structural neighbors, independently from whether message-passing is done in the forward or backward pass.
Isolating the inductive bias of a model from training is not completely unprecedented, as it was previously employed for the analysis of recurrent neural networks \cite{Tino2003,Tino2004,Gallicchio2011}.
For our experiments, we adopt two training-free models with different architectural biases, \textsl{Simplified Graph Convolution} (SGC) \cite{Wu2019} and \textsl{Graph Echo State Networks} (GESN) \cite{Gallicchio2010}.
In particular, the latter has achieved performances in line with or better than widely adopted end-to-end trained GNNs in node classification tasks \cite{Tortorella2023neurocomp}, also significantly improving upon previous results that include rewiring as graph pre-processing \cite{Tortorella2022log}.
This may suggest that the training process itself can pose serious challenges.

\begin{figure}
	\centering
	\begin{subfigure}{\columnwidth}
		\centering
		\includegraphics[width=.9\textwidth]{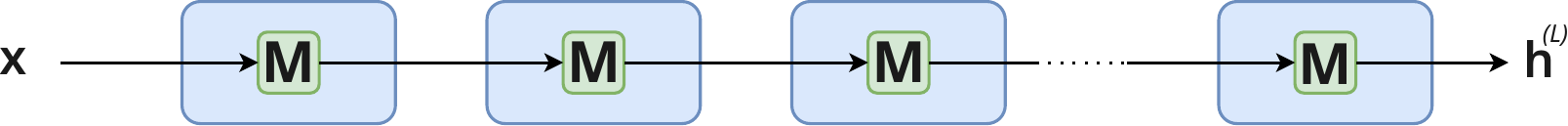}
		\subcaption{SGC}
		\label{fig:sgc}
	\end{subfigure}
	\begin{subfigure}{\columnwidth}
		\centering
		\includegraphics[width=.9\textwidth]{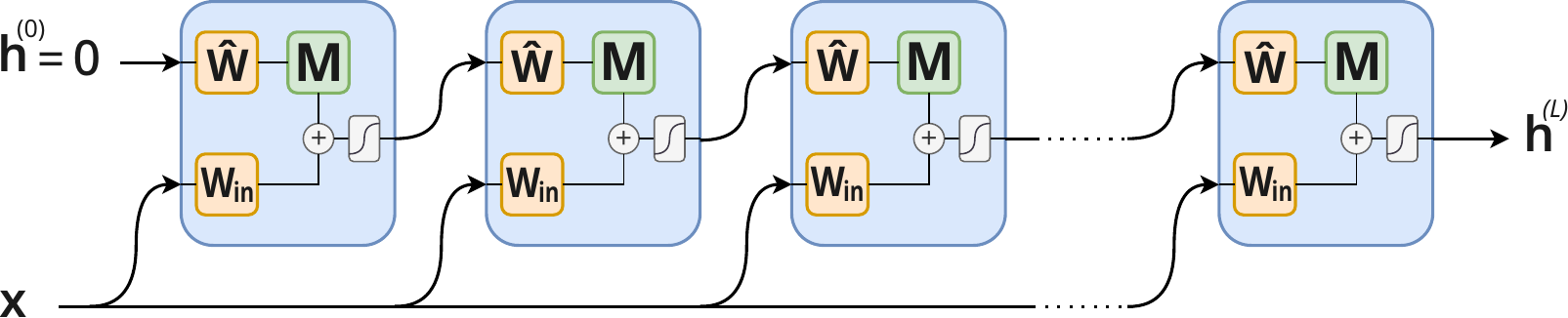}
		\subcaption{GESN}
		\label{fig:gesn}
	\end{subfigure}
	\caption{The two different model architectures of SGC \cite{Wu2019} and GESN \cite{Gallicchio2010}.}
	\label{fig:architectures}
\end{figure}

\subsection{Simplified graph convolution}
A straightforward way to compute node representations without the need for training is to replace the functions $\phi_k, \psi_k$ in \eqref{eq:mpnn} with the identity, thus removing altogether parameters in layers.
Such an approach was previously proposed by \cite{Wu2019} as a simplification of graph convolution by removing non-linearities, hence the name SGC.
This model therefore is reduced to pure message-passing (Fig.~\ref{fig:sgc}), with node representations computed after $L$ message-passing steps as
\begin{equation}\label{eq:sgc}\textstyle
	\vect{h}^{(L)} = \matx{M}^L\, \vect{x}.
\end{equation}
Notice that this model was proposed exclusively for node-level tasks \cite{Wu2019}.

\subsection{Graph echo state networks}
A different approach for training-free models is to follow the reservoir computing (RC) paradigm \cite{Nakajima2021,Lukosevicius2009,Verstraeten2007}, where input representations (or embeddings) are computed by a dynamical system with randomly initialized parameters.
Combining the recursive embedding approach of \cite{Scarselli2009} with RC, Graph Echo State Networks \cite{Gallicchio2010} compute node representations by iterating up to $L$ times the same message-passing function
\begin{equation}\label{eq:gesn}\textstyle
	\vect{h}_v^{(k)} = \tanh\left(\matx{W}_{\text{in}}\, \vect{x}_v + \sum_{u \in \mathcal{V}} M_{u v} \matx{\hat{W}}\, \vect{h}_{u}^{(k-1)} + \vect{b}\right),\quad \vect{h}_v^{(0)} = \vect{0},
\end{equation}
where $\matx{W}_{\text{in}} \in \mathbb{R}^{H \times X}$, $\vect{b} \in \mathbb{R}^H$ and $\matx{\hat{W}} \in \mathbb{R}^{H \times H}$ are respectively the input-to-reservoir, bias, and recurrent weights for a reservoir with $H$ units.
This can be interpreted as a form of parameter sharing between message-passing layers (Fig.~\ref{fig:gesn}).
Notice also that equation \eqref{eq:gesn} slightly departs from \eqref{eq:mpnn} due to the presence of input skip connections.
All reservoir weights are randomly initialized, with $\matx{W}_{\text{in}}$ rescaled to accommodate the input features' range, and $\matx{\hat{W}}$ rescaled to control the Lipschitz constant of \eqref{eq:gesn}.
For $\norm{\matx{\hat{W}}}\, \norm{\matx{M}} < 1$ the message-passing function is contractive \cite{Tortorella2022ijcnn}, that is the iterations of \eqref{eq:gesn} converge to a fixed point $\vect{h}^{(\infty)}$ as $L \to \infty$.
While this regime has been shown to be optimal for graph-level tasks, node-level tasks instead benefit from a non-contractive initialization $\norm{\matx{\hat{W}}}\, \norm{\matx{M}} > 1$, as the upper bound on input sensitivity \eqref{eq:sensitivity} intuitively suggests.
In the non-contractive regime, a choice of $L$ larger than the graph diameter is sufficient to ensure effective node representations \cite{Tortorella2023neurocomp}.

To produce graph representations for graph-level tasks, we apply a parameter-free global pooling operation, such as sum or mean pooling, to the final node representations:
\begin{equation}\label{eq:pooling}\textstyle
	\vect{h}_\mathcal{G}^\textsc{sum} = \sum_{v \in \mathcal{V}} \vect{h}_v^{(L)},\quad \vect{h}_\mathcal{G}^\textsc{mean} = \frac{1}{|\mathcal{V}|} \sum_{v \in \mathcal{V}} \vect{h}_v^{(L)}.
\end{equation}

\subsection{Readout classifier}
To solve a downstream node (or graph) classification task, there still remains the need to train a predictor.
For this purpose, we use a linear readout layer
\begin{equation} \label{eq:readout}
    \vect{\hat{y}}_v = \matx{W}_\text{out}\, \vect{h}_v^{(L)} + \vect{b}_\text{out},
\end{equation}
where the weights $\matx{W}_\mathrm{out} \in \mathbb{R}^{C \times H},\, \vect{b}_\mathrm{out} \in \mathbb{R}^C$ are trained by ridge regression on one-hot encodings of target classes $y_v \in 1, ..., C$.
In multi-class settings, the predicted class is obtained as $\argmax_{1 \leq c \leq C} \hat{y}_v^{(c)}$.
Training the classifier \eqref{eq:readout} can be achieved efficiently in closed form even on large data via the pseudo-inverse method, thus removing also from the readout any issue connected to back-propagation learning.
We refer to \cite{Zhang2017} for the details on how $\matx{W}_\text{out},\, \vect{b}_\text{out}$ are computed.

\section{Experiments and discussion}
\label{sec:exper}

We evaluate the graph rewiring methods of Sec.~\ref{sec:rewiring} jointly with the training-free GNNs presented in the previous section on several real-world classification tasks, many of whom were also adopted in previous rewiring literature \cite{Topping2022,ArnaizRodriguez2022}.
The aim of our experimental approach is to provide a tool for examining the effects of rewiring from a different perspective than previously pursued in the literature, thanks to decoupling the inductive bias of GNNs from the training process.

\subsection{Datasets}

\begin{table*}[]
    \caption{Statistics of the datasets adopted in our experiments.}
    \label{tab:datasets}
    \begingroup
	\footnotesize
	\setlength{\tabcolsep}{2.5pt}
    \begin{subtable}{.5\textwidth}
	\begin{tabular}{lrrrrrr}
		\toprule
		\multicolumn{7}{c}{\textsc{Node Classification}} \\
		\midrule
		& \textbf{Cora} & \textbf{CiteSeer} & \textbf{PubMed} & \textbf{Film} & \textbf{TwitchDE} & \textbf{Tolokers} \\
		\midrule
		nodes          & $2\mathord{,}708$ & $3\mathord{,}327$ & $19\mathord{,}717$ & $7\mathord{,}600$ & $9\mathord{,}498$ & $11\mathord{,}758$ \\
		edges          & $10\mathord{,}556$ & $9\mathord{,}104$ & $88\mathord{,}648$ & $53\mathord{,}504$ & $153\mathord{,}138$ & $519\mathord{,}000$ \\
		avg. degree & $3.90$ & $2.74$ & $4.50$ & $7.03$ & $16.14$ & $88.28$ \\
		diameter       & $19$ & $28$ & $18$ & $12$ & $7$ & $11$ \\
		node features  & $1\mathord{,}433$ & $3\mathord{,}703$ & $500$ & $932$ & $2\mathord{,}514$ & $10$ \\
		classes        & $7$ & $6$ & $3$ & $5$ & $2$ & $2$ \\
		edge homophily & $0.81$ & $0.74$ & $0.80$ & $0.22$ & $0.63$ & $0.59$ \\
		\bottomrule
	\end{tabular}
    \end{subtable}
    \begin{subtable}{.4\textwidth}
	\begin{tabular}{lrrrrrr}
		\toprule
		\multicolumn{7}{c}{\textsc{Graph Classification}} \\
		\midrule
		& \textbf{NCI-1} & \textbf{NCI-109} & \textbf{Reddit-B} & \textbf{Reddit-5K} & \textbf{Reddit-12K} & \textbf{Collab} \\
		\midrule
		graphs           & $4\mathord{,}110$ & $4\mathord{,}127$ & $2\mathord{,}000$ & $4\mathord{,}999$ & $11\mathord{,}929$ & $5\mathord{,}000$ \\
		avg. nodes    & $30$ & $30$ & $430$ & $509$ & $391$ & $75$ \\
		avg. edges    & $32$ & $32$ & $498$ & $595$ & $457$ & $2\mathord{,}458$ \\
		avg. degree   & $2.16$ & $2.16$ & $2.34$ & $2.25$ & $2.28$ & $37.37$ \\
		avg. diameter & $13.33$ & $13.13$ & $9.72$ & $11.96$ & $10.91$ & $1.86$ \\
		node features    & $37$ & $38$ & $0$ & $0$ & $0$ & $0$ \\
		classes          & $2$ & $2$ & $2$ & $5$ & $11$ & $3$ \\
		\bottomrule
	\end{tabular}
    \end{subtable}
    \endgroup
\end{table*}

For node classification tasks, we adopt six graphs of up to $20\mathord{,}000$ nodes.
Cora \cite{McCallum2000}, CiteSeer \cite{Giles1998}, PubMed \cite{Sen2008} are paper citation networks, where input node features are bag-of-words representations of paper content, and the target is the research topic.
Film \cite{Tang2009} is a network induced by co-occurrences of actors in the same Wikipedia page, grouped into five categories \cite{Pei2020}.
TwitchDE \cite{Rozemberczki2021,Lim2021} is a social network of German gamer accounts from Twitch classified into suitable for work or adult profiles.
Tolokers \cite{Likhobaba2023,Platonov2023} is a collaboration network of users extracted from the crowdsourcing platform Toloka, where the task is to determine whether a user is active or not (since the two classes are unbalanced, the evaluation metric in this case is area under the ROC curve instead of accuracy).
The first three are homophilous node classification tasks, while the other three tasks present low homophily.
For graph classification, we adopt six tasks from the TUDataset collection \cite{Morris2020}.
NCI-1, NCI-109 \cite{Wale2007,Shervashidze2011} are molecules to be classified as cancerogenous or not, where node input features are one-hot encodings of atom type, and edges correspond to chemical bounds.
Reddit-B, Reddit-5K, Reddit-12K \cite{Yanardag2015} are interaction networks between users in Reddit discussion threads, and the classification task is to identify the type of sub-reddit the discussions belong to.
Collab \cite{Leskovec2005,Yanardag2015} is a collection of ego-networks belonging to three different scientific collaboration fields.
Both Reddit tasks and Collab have no node input features.
In all tasks, we have consciously avoided adding structural input features to the graph nodes, such as node degrees or positional encodings \cite{Srinivasan2020}.
Relevant dataset statistics are reported in Tab.~\ref{tab:datasets}.

\subsection{Experimental setting}
For all classification tasks we have generated with class stratification $5$-fold selection/test splits with inner validation holdout, thus resulting in 60:20:20 training/validation/test set ratios.
Both GNN and rewiring algorithm parameters are jointly selected on each validation fold.
For SGC, we select the number of message-passing iterations $L \in [1, 15]$ and the type of message-passing matrix (adjacency, Laplacian, or one of their normalizations, with or without the addition of self-loops).
For GESN, we select the reservoir size (i.e. node representation dimension) $H \in [2^4, 2^{12}]$, the input scaling factor in $[0, 1]$, and the Lipschitz constant.
For the latter, we actually follow the reservoir computing practice of selecting the spectral radius $\rho(\matx{\hat{W}})$ instead of the spectral norm $\norm{\matx{\hat{W}}}$, as the radius is a lower bound on the norm \cite{Goldberg1974} and it is cheaper to compute \cite{Gallicchio2020inns}.
We select $\rho(\matx{\hat{W}}) \in [0.1/\rho(\matx{M}), 30/\rho(\matx{M})]$, while the number of message-passing iterations is fixed at $L = 30$, which is comfortably larger than graph diameters in our datasets \cite{Tortorella2023neurocomp}.
For graph-level tasks we also select the pooling function from the two defined in \eqref{eq:pooling}.
As for graph rewiring algorithms, we select $t \in [0.1, 5]$ for heat diffusion, and $\alpha \in [0.01, 0.99]$ for PageRank diffusion.
We run SDRF and GRLEF for a number of iterations corresponding to up to $20\%$ of the graph edges, without performing edge removal in the former.
Finally, the regularization for the closed-form ridge regression to train the readout classifier is selected in $[10^{-5},10^3]$.

\begin{table}[t]
	\centering
	\caption{Node classification with SGC.}
	\label{tab:node-sgc}
	\begingroup
	\footnotesize
	\setlength{\tabcolsep}{1.5pt}
	\begin{tabular}{lcccccc}
		\toprule
		& \textbf{Cora} & \textbf{CiteSeer} & \textbf{PubMed} & \textbf{Film} & \textbf{TwitchDE} & \textbf{Tolokers} \\
		\midrule
		\textsl{Baseline} & $87.81_{\pm 2.00}$ & $76.86_{\pm 1.07}$ & $87.98_{\pm 0.43}$ & $32.08_{\pm 0.53}$ & $67.11_{\pm 1.02}$ & $74.77_{\pm 1.50}$ \\
		\cmidrule{2-7}
		Heat     & $87.41_{\pm 2.07}$ & ${76.89}_{\pm 0.92}$ & \better{$88.73_{\pm 0.44}$} & \better{$33.88_{\pm 1.43}$} & ${67.78}_{\pm 0.74}$ & $75.82_{\pm 0.70}$ \\
		PageRank & ${87.85}_{\pm 2.03}$ & $76.28_{\pm 0.81}$ & \better{${88.83}_{\pm 0.51}$} & \better{${34.86}_{\pm 1.75}$} & $67.00_{\pm 1.15}$ & ${75.88}_{\pm 0.63}$ \\
		SDRF     & $86.78_{\pm 1.83}$ & $76.65_{\pm 1.41}$ & \worse{$87.38_{\pm 0.37}$} & $32.01_{\pm 1.29}$ & $67.62_{\pm 0.50}$ & OOR              \\
		GRLEF    & \worse{$85.08_{\pm 1.98}$} & $75.47_{\pm 2.09}$ & \worse{$87.02_{\pm 0.45}$} & $31.41_{\pm 1.31}$ & \worse{$65.90_{\pm 0.43}$} & $72.96_{\pm 1.91}$ \\
		EGP      & $87.81_{\pm 2.00}$ & $76.86_{\pm 1.07}$ & $87.98_{\pm 0.43}$ & \worse{$29.25_{\pm 1.37}$} & $67.11_{\pm 1.02}$ & $74.77_{\pm 1.50}$ \\
		DiffWire & \worse{$84.68_{\pm 1.51}$} & \worse{$73.43_{\pm 1.44}$} & \worse{$84.73_{\pm 0.22}$} & $31.46_{\pm 0.90}$ & $67.37_{\pm 0.32}$ & $75.27_{\pm 1.34}$ \\
		\bottomrule
	\end{tabular}
	\endgroup
\end{table}

\begin{table}
	\centering
	\caption{Node classification with GESN.}
	\label{tab:node-gesn}
	\begingroup
	\footnotesize
	\setlength{\tabcolsep}{1.5pt}
	\begin{tabular}{lcccccc}
		\toprule
		& \textbf{Cora} & \textbf{CiteSeer} & \textbf{PubMed} & \textbf{Film} & \textbf{TwitchDE} & \textbf{Tolokers} \\
		\midrule
		\textsl{Baseline} & $87.70_{\pm 1.34}$ & ${75.84}_{\pm 0.93}$ & ${89.53}_{\pm 0.49}$ & $35.23_{\pm 0.70}$ & $68.62_{\pm 1.04}$ & $84.40_{\pm 1.02}$ \\
		\cmidrule{2-7}
		Heat     & ${87.86}_{\pm 1.50}$ & $75.34_{\pm 0.88}$ & $89.22_{\pm 0.33}$ & \better{${36.87}_{\pm 1.05}$} & $68.26_{\pm 0.30}$ & $84.20_{\pm 1.17}$ \\
		PageRank & $87.50_{\pm 1.30}$ & $75.20_{\pm 1.32}$ & $89.19_{\pm 0.42}$ & $35.91_{\pm 1.06}$ & $67.88_{\pm 0.49}$ & \worse{$82.63_{\pm 1.18}$} \\
		SDRF     & $86.60_{\pm 1.56}$ & $74.84_{\pm 1.66}$ & $89.20_{\pm 0.40}$ & $34.92_{\pm 0.55}$ & $68.54_{\pm 0.80}$ &  OOR             \\
		GRLEF    & $86.06_{\pm 1.56}$ & $74.74_{\pm 1.73}$ & $89.11_{\pm 0.74}$ & $35.05_{\pm 0.87}$ & $67.66_{\pm 0.70}$ & \worse{$82.64_{\pm 1.19}$} \\
		EGP      & $86.95_{\pm 2.51}$ & $74.62_{\pm 1.85}$ & $89.50_{\pm 0.42}$ & $35.06_{\pm 0.78}$ & ${68.68}_{\pm 0.98}$ & $84.50_{\pm 1.02}$ \\
		DiffWire & $86.51_{\pm 1.74}$ & $74.03_{\pm 2.20}$ & \worse{$88.81_{\pm 0.49}$} & $35.01_{\pm 0.74}$ & $68.15_{\pm 0.33}$ & ${84.77}_{\pm 0.95}$ \\
		\bottomrule
	\end{tabular}
	\endgroup
\end{table}

\begin{table}
	\centering
	\caption{Graph classification with GESN.}
	\label{tab:graph-gesn}
	\begingroup
	\footnotesize
	\setlength{\tabcolsep}{1.5pt}
	\begin{tabular}{lcccccc}
		\toprule
		& \textbf{NCI-1} & \textbf{NCI-109} & \textbf{Reddit-B} & \textbf{Reddit-5K} & \textbf{Reddit-12K} & \textbf{Collab} \\
		\midrule
		\textsl{Baseline} & $78.09_{\pm 1.64}$ & $77.56_{\pm 0.83}$ & $87.23_{\pm 1.38}$ & $53.86_{\pm 1.49}$ & $44.02_{\pm 0.54}$ & $72.49_{\pm 0.77}$ \\
		\cmidrule{2-7}
		SDRF      & \worse{$73.39_{\pm 0.63}$} & \worse{$72.35_{\pm 1.46}$} & $87.02_{\pm 1.30}$ & $53.84_{\pm 1.55}$ & $44.07_{\pm 0.47}$ & $71.25_{\pm 1.09}$ \\
		GRLEF     & \worse{$73.74_{\pm 1.40}$} & \worse{$71.76_{\pm 1.31}$} & $85.89_{\pm 2.02}$ & $53.17_{\pm 1.26}$ & $42.94_{\pm 1.23}$ & $72.23_{\pm 0.86}$ \\
		EGP       & $78.31_{\pm 1.63}$ & $77.49_{\pm 0.65}$ & $87.28_{\pm 1.29}$ & $53.78_{\pm 1.34}$ & $44.08_{\pm 0.48}$ & $72.17_{\pm 0.87}$ \\
		DiffWire  & $78.14_{\pm 1.61}$ & $77.48_{\pm 0.65}$ & $84.54_{\pm 2.44}$ & $53.58_{\pm 0.72}$ & \worse{$41.37_{\pm 0.68}$} & \worse{$66.31_{\pm 0.76}$} \\
		\bottomrule
	\end{tabular}
	\endgroup
\end{table}

\subsection{Discussion of results}
We report the results of our experiments in Tab.~\ref{tab:node-sgc}--\ref{tab:graph-gesn}.
The baseline accuracy corresponds to the model applied to the original graph without any rewiring.
We have applied the one-sided $t$-test with a significance level of $p < 0.05$ to highlight whether the classification accuracy achieved by a rewiring method is either \better{better} or \worse{worse} compared to the respective baseline value without rewiring, denoting no significant difference otherwise.
The experiments were executed on an NVIDIA A100 with 40GB of GPU RAM.
For reference, a single complete model selection for GESN excluding rewiring took up to $3.5$ hours.
`OOR' in Tab.~\ref{tab:node-sgc}--\ref{tab:node-gesn} indicates that SDRF exceeded the limit of 10 days of computation for Tolokers.

\begin{figure}
	\centering
	\begin{subfigure}{\columnwidth}
		\includegraphics[width=\textwidth]{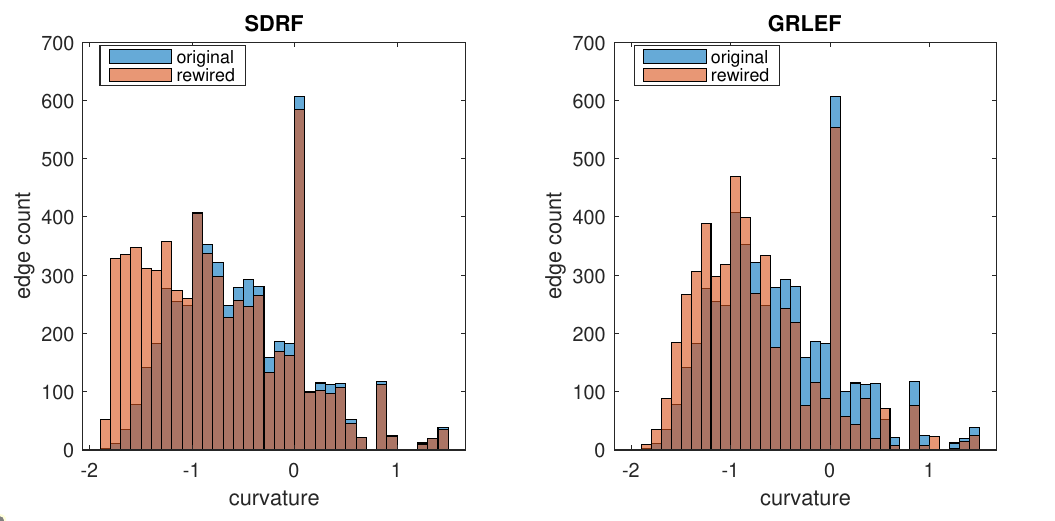}
		\caption{Edge curvature distribution}
		\label{fig:cora-histogram}
	\end{subfigure}
	\begin{subfigure}{\columnwidth}
		\includegraphics[width=\textwidth]{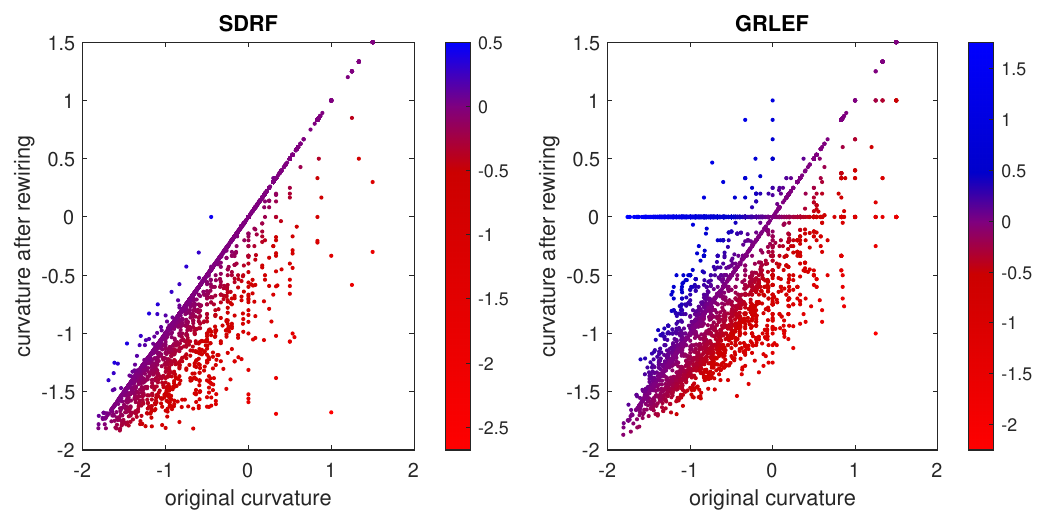}
		\caption{Edge curvature variation}
		\label{fig:cora-deltas}
	\end{subfigure}
	\caption{Effects of SDRF and GRLEF on graph curvature $\mathsf{Ric}$ for Cora.}
	\label{fig:cora-curvatures}
\end{figure}

On node classification tasks, the only rewiring methods able to achieve some significant improvements over the baseline both for SGC and GESN are the diffusion-based heat and PageRank rewiring.
This improvement is present both on high and low homophily graphs, that is, respectively PubMed and Film.
This comes as a surprise, since these methods were dismissed in previous literature \cite{Topping2022}.
We may conjecture that by acting as low-pass filters on the graph spectra \cite{Klicpera2019}, diffusion methods can improve the spectral gap $\lambda_1$ (i.e. the smallest positive Laplacian eigenvalue) in certain graphs, possibly resulting in a rewired graph with a larger Cheeger constant since $\mathfrak{h}_\mathcal{G} \geq \frac{\lambda_1}{2}$ \cite{Mohar1989}.
The other rewiring methods do not provide significant improvements in accuracy, both on node and graph classification tasks.
Actually, they can cause a significant degradation of accuracy.
To investigate the effects of rewiring algorithms that explicitly act on local bottlenecks of graph topology, we analyze the distribution of edge curvature before and after rewiring (Fig.~\ref{fig:cora-histogram}).
As stated in Sec.~\ref{sec:sdrf}, a prevalence of positively-curved edges would denote a polynomial rather than exponential receptive field growth.
Notice that the overall curvature distribution is not improved; in particular, the one of SDRF appears to become even more skewed towards negatively curved edges. 
This is confirmed by observing the differences between initial and final edge curvature in the scatter plots of Fig.~\ref{fig:cora-deltas}, where the predominant number of edges appears in red below the diagonal, denoting that edge curvature has actually become more negative instead of improving.
This behavior can be explained by recalling that the algorithm acts \emph{greedily} on \emph{local} curvature information, without accounting for the effects on \emph{global} graph curvature when deciding where to add supporting edges \cite{DiGiovanni2023}.
\begin{figure}
	\centering
	\includegraphics[width=.65\columnwidth]{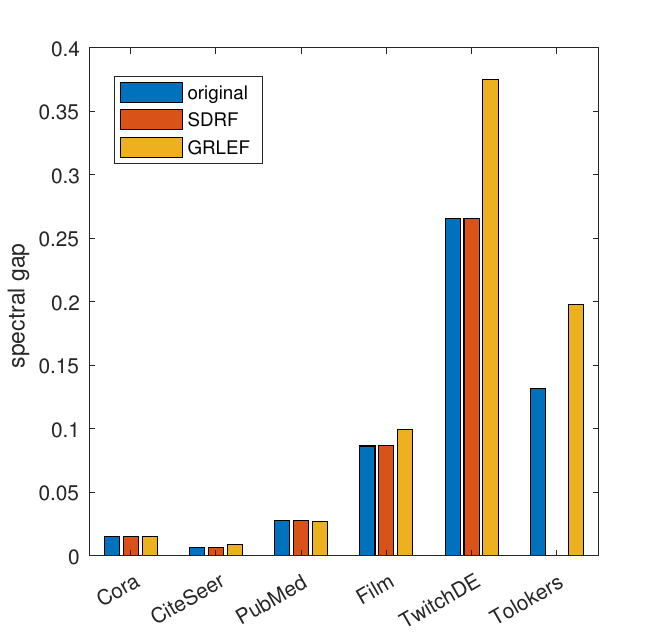}
	\caption{Effects of SDRF and GRLEF on spectral gaps $\lambda_1$.}
	\label{fig:gaps}
\end{figure}
As stated in Sec.~\ref{sec:grlef}, the spectral gap is a proxy measure of global graph bottlenecks.
In Fig.~\ref{fig:gaps} we analyze the effects of the two local rewiring algorithms SDRF and GRLEF on this global property.
While a more positive curvature should also improve the spectral gap since $\lambda_1 \geq \min_{(u,v) \in \mathcal{E}} \mathsf{Ric}_{u v}$ \cite{Topping2022}, the failure of SDRF in generally increasing edge curvature results in an unchanged $\lambda_1$.
On the other hand, GRLEF is in some cases able to provide some increase in the spectral gap.
However, this does not necessarily translates into an improvement of node classification accuracy, as the results on TwitchDE and Tolokers show.
EGP seems to have little effect on accuracy both on graph and node classification tasks in general.
As for DiffWire, the significant degradation of accuracy on Collab and Reddit-12K could be attributed to a magnification of spurious edges between network communities.
While the scope of our empirical approach is to validate the effectiveness of graph rewiring purely on message-passing, to put the results of our experiments into perspective, we recall that the addition of training to a message-passing model on the rewired graph has shown no improvements over training-free baselines \cite{Tortorella2022log}.

\subsection{Relation with prior literature results}

The results of \cite{Topping2022} have been recently questioned by \citet{Tori2025}, which have conducted a reproducibility reassessment of the accuracy improvements of curvature-based rewiring approaches in \emph{fully-trained} GNNs.
When much broader hyperparameter ranges are explored in the model selection of GNNs, no statistically significant accuracy shift has been found when SDRF-like rewiring methods are applied jointly with the same trained GNNs.
The results of the remaining rewiring algorithms in the respective original papers should also be put into perspective.
In \cite{Banerjee2022}, GRLEF has not been assessed on real-world node or graph classification tasks.
In \cite{Deac2022}, EGP has been tested on a single trained GNN model only in a few graph classification benchmarks, without any statistical significance analysis.
As DiffWire in \cite{ArnaizRodriguez2022} \emph{learns} the effective resistance jointly in the GNN training instead of computing it, their experiments cannot conclusively assess the benefits of resistance-based rewiring.
On the other hand, our results for diffusion-based methods are consistent with the edge denoising effects stated in \cite{Klicpera2019}.

\subsection{Computational overhead of rewiring}

\begin{table}
	\centering
	\caption{Overhead cost of rewiring in terms of computational complexity. $V$ and $E$ denote the number of nodes and edges, while $d_\text{max}$ is the maximum degree.}
	\label{tab:complexity}
	\begingroup
	\footnotesize
	\begin{tabular}{ll}
		\toprule
		\textbf{Rewiring} & \textbf{Computational complexity} \\
		\midrule
		Heat     & $O(V^3)$ for closed-form computation of $\matx{M}_\text{Heat}$ \\
		PageRank & $O(V^3)$ for closed-form computation of $\matx{M}_\text{PageRank}$ \\
		SDRF     & $O(V^2 E d_\text{max}^2)$ for each new edge insertion \\
		GRLEF    & $O(E d_\text{max}^2)$ for each edge flip \\
		EGP      & \makecell[l]{$O(E')$ for each additional message-passing step \\ on an expander graph with $E'$ edges} \\
		DiffWire & $O(V^3)$ for closed-form computation of $\matx{M}_\text{DiffWire}$ \\
		\bottomrule
	\end{tabular}
	\endgroup
\end{table}

The computational complexity of message-passing depends on the functions $\phi_k, \psi_k$ in eq.~\eqref{eq:mpnn}: for example, bare neighborhood aggregation is $O(E H)$ in SGC, while for GESN the complexity is  $O(E H + V H^2)$.
Except for EGP, rewiring algorithms constitute an overhead cost paid during graph pre-processing that can be extremely demanding (Tab.~\ref{tab:complexity}).
The computation of diffusion and effective resistance matrices involves matrix inversion, which in standard form is $O(V^3)$, while SDRF and GRLEF require multiple computations of curvature measures.

\subsection{Limitations}
Our evaluation has focused on rewiring methods that are applied as a pre-processing step to GNNs, whereas some authors consider models that perform structure learning or attention mechanisms as `implicit' rewiring approaches.
We may also have left out from our evaluation rewiring methods that were proposed only very recently or that do not have a publicly available reference implementation.
For our experiments, we have chosen well-used, real-world classification tasks.
While artificially constructed tasks can be useful to compare the ability of rewiring methods to deal with particular graph topologies, we have chosen to focus on situations that may arise in practical applications.
Finally, in our tasks we have considered only undirected and unweighted graphs.
This limitation is shared with some rewiring methods we have analyzed: the curvature measure on which SDRF and GRLEF are based is itself defined only for undirected and unweighted edges.

Keeping in line with prior literature, we have focused on investigating the effects of rewiring graph topology in the context of classification tasks via GNNs.
At the best of our knowledge, no study has been conducted on the effects of rewiring in the contexts of other tasks such as graph partitioning and clustering.
Graph diffusion methods have been applied to such tasks (e.g. \cite{Andersen2007,KalalaMutombo2024}).
In general, we could only conjecture that local rewiring methods may have less severe effects than methods which create long-range bridges between nodes.

\section{Conclusions}
\label{sec:concl}

The study of limits of message-passing in graph neural networks is becoming an increasingly vibrant research field, with both theoretical analysis on the factors that cause over-squashing and solutions based on altering the input graph being proposed \cite{Topping2022,DiGiovanni2023,Black2023}.
In this paper, we have attempted to answer to the need for a rigorous experimental evaluation setting to assess the benefits of graph rewiring methods.
We have proposed the use of message-passing models that compute node and graph representations without training, in order to decouple the issues connected with over-squashing from those derived from the training process, such as gradient vanishing.
Indeed, such models have previously achieved performances in line with or better than most end-to-end trained GNNs, e.g. on node classification \cite{Tortorella2023neurocomp}.
The results on twelve real-world node and graph classification tasks have shown that in most cases, rewiring does not offer significant practical benefits for supporting message-passing.
As the GESN baseline has also outperformed trained GNNs with rewiring \cite{Tortorella2022log}, this may possibly suggest that the problems that GNNs have to face are more connected with how training is conducted than with obstacles in message-passing derived from graph topology.
In this paper, we have thus also offered training-free baseline models against which to test the practical advantages of other GNNs with end-to-end training.
Moreover, whenever the computational cost of rewiring pre-processing or just simply training GNNs is significant, the same training-free models may offer a feasible and effective alternative.

\section*{Acknowledgments}
Research partly supported by:
PNRR, PE00000013, ``FAIR - Future Artificial Intelligence Research'', Spoke 1, funded by European Commission under NextGeneration EU programme;
Project DEEP-GRAPH, funded by the Italian Ministry of University and Research, PRIN 2022 (project code: 2022YLRBTT, CUP: I53C24002440006);
Project PAN-HUB, funded by the Italian Ministry of Health (POS 2014--2020, project ID: T4-AN-07, CUP: I53C22001300001).

\bibliographystyle{elsarticle-num-names}
\bibliography{bibliography}



\end{document}